# Towards Long-Distance Inspection for In-pipe Robots in Water Distribution Systems with Smart Navigation

**Saber Kazeminasab [1,*] and M Katherine Banks [2]**

[1] Department of Electrical and Computer Engineering, Texas A&M University, Texas, USA; skazeminasab@tamu.edu

[2] College of Engineering, Texas A&M University, Texas, USA; k-banks@tamu.edu

**Abstract:** Incident in water distribution systems (WDS) cause water loss and water contamination that requires the utility managers to assess the condition of pipelines in a timely manner. However, pipelines are long and access to all parts of it is a challenging task; current in-pipe robots have the limitations of short-distance inspection and inability to operate in-service networks. In this work, we improve the design of our previously developed in-pipe robot and analyze the effect of line pressure and relative velocity on the robot during operation with computational fluid dynamics (CFD) simulations. An extreme scenario for robot operation is defined and we estimate the minimum inspection distance for the robot with one turn of battery charge that is 5400m. A multi-phase motion controller is proposed that ensures reliable motion at straight and non-straight configurations of pipeline and also stabilized configuration with zero velocity at junctions. We also propose a localization and navigation method based on particle filtering and combine it with the proposed multi-phase motion controller. In this method, the map of the operation is provided to the robot and the robot localizes itself based on the map and the particle filtering method. Furthermore, the robot navigates different configurations of pipelines by switching between different phases of the motion controller algorithm that is performed by the particle filter algorithm. The experiment and simulation results show that the robot along with the navigation shows a promising solution towards long-distance inspection of pipelines by in-pipe robots.

**Keywords**: In-pipe Robots; Water Distribution Networks; Long-distance Inspection, Smart Navigation.

## 1. Introduction

Water distribution systems (WDS) are one of the strategic infrastructures that carry potable water in millions of miles of lines around the globe. These pipelines are prone to corrosion, damages, and even incidents in the network that cause leaks and water loss. Around 240,000 water main breaks occur in the U.S. annually which are responsible for 2 trillion gallons loss [1] and account for 15%-25% of the purified water [2]. The water loss accounts for 20% of purified water in Canada [3] and hence, it is required to assess the condition of pipelines periodically, and in case of a leak, the utilities need to localize and repair it [4] or monitor the quality of water [5,6]. To this aim, mobile sensors are proposed for in-pipe operation [7]. These mobile sensors move in-pipes with passive motion and perform the desired task. However, due to passiveness, the operator may lose them in the network [8], and they cannot inspect long distances of pipelines. To address passiveness, in-pipe robots are designed in which their motion is independent of flow. In the literature, the robots that are in contact with pipe wall are common mechanisms [9–13] while free-swimming robots are also designed for small pipes [14]. Several factors impact the motion of these robots; the operation condition of these robots are pipelines with varying sizes in which high-pressure flow is present that applies huge disturbances on the robot [15], the robot path is not always a horizontal path and the inclination angle of the pipe affects the motion of the robot. Besides, pipeline maps are not accurate [16], and due to sediments in pipes [17], the internal shape of pipes is not known and there are many uncertainties in pipelines that require the robotic systems to be stable and robust to these disturbances and uncertainties [18]. Space limitation is another deciding factor that affects the size of the robot and it is highly desired that the robot operates in the pipelines while the network remains in service. Under-actuated mechanical systems [19] have shown promising capabilities for tackling uncertain environments with a proper motion controller [20]. The number of control inputs is fewer than the degrees of freedom (DoF) in these robots and hence, stabilization is the prime factor that needs to be fulfilled and normally a stabilizer controller is combined with a trajectory controller to acquire a desired and stabilized motion [21–26]. They can be used in the design of in-pipe robots to benefit from their rather simple mechanism (compared to fully-actuated systems) and make the robot compact and maneuverable.

Long-distance inspection with smart navigation is desirable for in-pipe robots and affects the applicability of the in-pipe robots. The challenge relies on simultaneous localization and mapping (SLAM) where the robot needs to localize itself for map generation, while for localization, the map of the environment is required. Hence, simultaneous localization and mapping are required to acquire optimal performance.

For underwater systems, using sonar images this aim is a common approach, however, due to acoustic waves, sonar images are low detectable and results in the problems like instability and reflection based on the roughness of the reflector surface. The authors in [27], implemented sonar-based localization to generate a coherent map of the unknown environment. Also, to estimate the vehicle position in the network, they used the particle filtering method. Lee et al. improved the accuracy of the sonar image using two probability-based methods that are the combination of particle filtering and Bayesian methods [28]. Navigation is another challenge for in-pipe robots in which the robot needs to travel in the environment of pipelines. Yuan et al. employed the augmented extended Kalman Filter (AEKF) in SLAM to address the issue [29]. They deployed a recursive approach to have an estimation of the parameters of the state and storing the poses and landmarks of the robot in a single state vector to build an accurate map. Authors in [30] proposed four different Bayes filter-based algorithms to better address the challenges of SLAM and navigation that are: The Extended Kalman Filter (EKF), Unscented Kalman Filter (UKF), Particle Filter (PF), and Marginalized Particle Filter (MPF) and evaluated each method an provided the advantageous and disadvantageous of each method.

However, SLAM for in-pipe robots is not well-addressed in the literature [31]. Landmark-based localization is a common method in localization for these types of robots. For example, branches and elbows are distinguishable in this method in the pipelines. The authors in [32,33] proposed a method for localization and mapping by combining landmark and pose information from sensors. In this method, a line laser beam is projected on the internal surface of the pipe wall to generate the map. In another work, Lee et al. presented a method for in-pipe navigation with landmark recognition using a designed illuminator [34] in which analyzes the images and computes the current position of the robot and the pipeline map. Time-of-flight (TOF) camera is a useful tool in which it provides 3D imagery that is used navigation, mapping, and landmark detection for navigation and landmark detection. The authors in [35] propose implemented the TOF camera for localization and landmark detection that feature extraction is accomplished by fitting a cylinder to images of the pipeline.

Also, camera-based methods attracted the researchers in this field for inspection challenges in in-pipe robots. Vision sensors such as Fish-eye and Omni-directional cameras have shown promising results [36]. The authors in [36] proposed a method for the self-localization challenge of an earthworm robot by integrating an omnidirectional rangefinder. In this method, an omni-directional camera and an omni-directional laser rangefinder are used to construct a piping shape. A size-adaptable in-pipe is developed that localizes itself and detects map rust on the pipe based on per-pixel classification with image processing in [37]. A checkpoint method is used as a guide for the robot to rust mapping. To increase the accuracy of localization of the pipeline inspection gauge (PIG), two approaches are proposed [38]; the Inertial navigation system (INS) dynamic model and 3-D reduced inertial sensor system (RISS).

As for predicting the direction and also the cornering in in-pipe robots, the whisker-like sensor has shown potential. The authors in [39] developed a bespoke sensor based on a whisker-like sensor that is used for elbow detection for smart navigation in an unknown pipeline network. Acoustic signal with pose-graph optimization is a solution In to improve the estimation of a robot's trajectory [40]. The authors proposed different methods for pose graph construction using hydrophone acoustic signal [40].

Featureless pipes make the SLAM for in-pipe robots challenging. Systematic noise and observation noise can result in an inaccurate judgment for the mobile robot and several works have been done to reduce this effect. KF (Kalman Filter) is an attractive method in Gaussian noise canceling in the linear systems and Extended Kalman Filter (EKF) in nonlinear systems that are linearized with Taylor expansion. However, Kalman filters are limited to noise-canceling with Gaussian distributions. The particle filter (PF) method addresses the issue and can compute the posterior probability of the states that are disturbed with noises with none Gaussian distributions [41]. Ke Ma et all. Proposed a method for SLAM in metal water pipes that implements RaoBlackwellised particle filter (RBPF) [31]. This method builds a map of pipe using vibration amplitude over space by a hydrophone. Then, the generated map in the SLAM algorithm. Also, YuPei Yan et all. proposed PF-SLAM algorithm by combining the particle filter and FastSLAM algorithm to accomplish the precise moving direction and also navigation of the in-pipe robot [42]. They also proposed a visual simultaneous localization and mapping (VSLAM) method for localization and mapping by integrating the information of multiple CCD cameras, an inertial navigation sensor, and ultrasonic rangefinders sensors.

### 1.1. *Technical Gap*

Localization and navigation of the in-pipe robots in complicated configurations of the pipelines is challenge, especially for metal pipes, since wireless communication methods is not practical in these environments; radio signals are faded in metal environments. Also, long-distance inspection with smart navigation in in-service distribution systems is highly desired and not well addressed in the literature.

### 1.2. *Our Contribution*

- We improve the design of our previously proposed in-pipe robot [43] and prototype it.

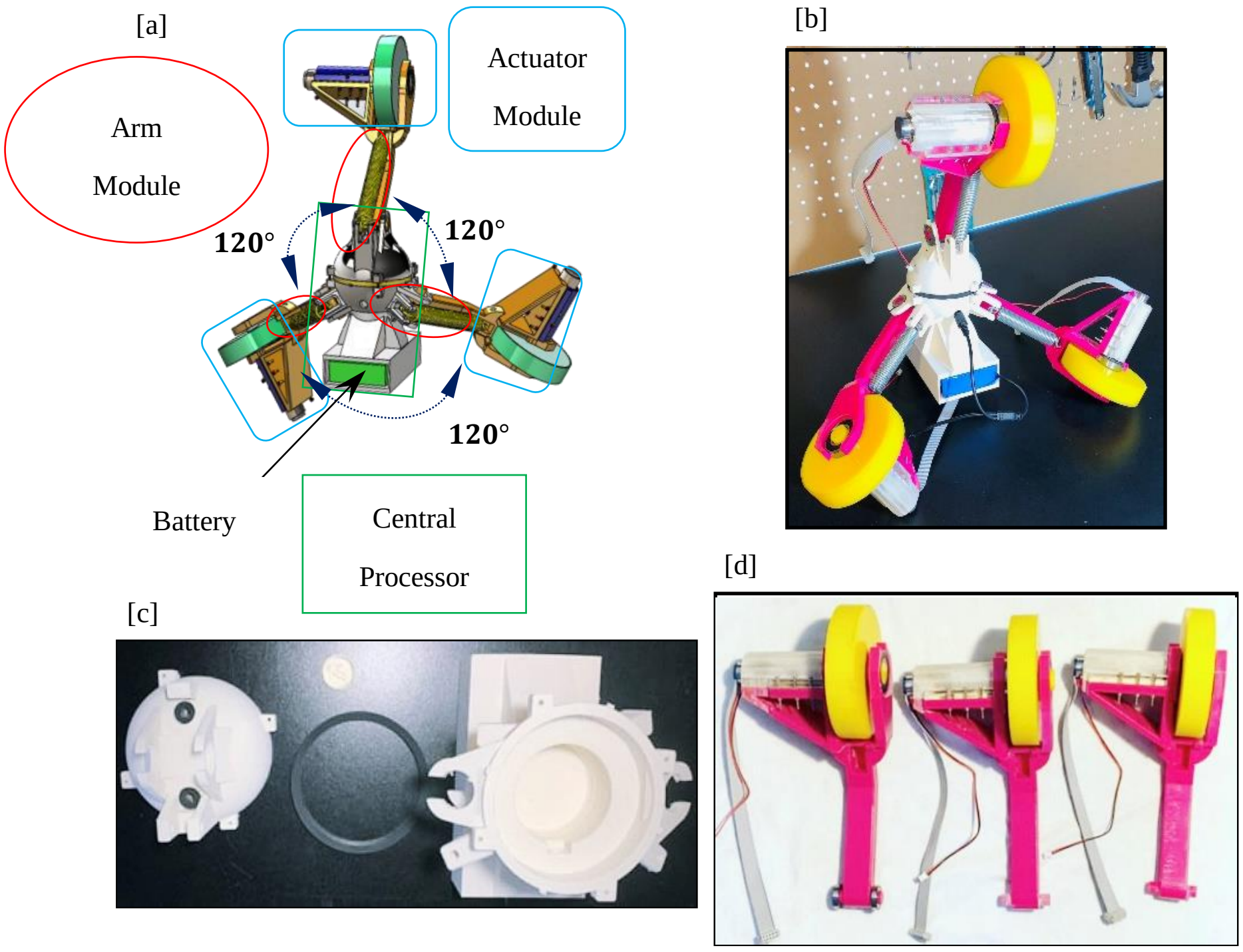

**Fig 1**. [a] CAD Design of the Proposed Robotic System. [b] Prototype of the Robot. [c] Prototyoe of the Central Processor. [d] Arm Module Equipped with Actuator Module.

- We analyze the parameters in the operating environment in in-service distribution networks that affect the performance of the robot during operation.
- Considering the affecting parameters, an analysis is provided that estimates the inspection distance of the robot in an extreme operating condition.
- We propose three phases of a motion control algorithm that ensures a reliable motion for the robot in complicated configurations of pipelines and also we validate their performance.
- We propose a localization method for the robot that enables it to localize itself in the network during operation.
- We propose a navigation method that enables the robot to navigate different configurations of the pipeline.

The remainder of the paper is organized as follows. In section 2, the robotic system is described. In section 3, the parameters that affect the performance of the robot are presented and analyzed. In section 4, the multi-phase control algorithm is presented and the performance of each phase is investigated. In section 5, the proposed localization and navigation method based on the combination of particle filtering method and multi-phase motion control algorithm is proposed and its performance is validated. The paper is then concluded in section 6.

## 2. The Robotic System Design

The robot comprises a central processor and three adjustable arms. The central processor performs sensor measurements and processes them, controls the motion of the robot, and communicates wirelessly. It includes two sphere-shaped components that locate the electronic components of the robot and a compact battery. The electronic parts are located in a printed circuit board (PCB) that is equipped with a seal mechanism to isolate it from the external environment. The battery provides power for different components of the robot during operation. The adjustable arms are anchored to the central processor with 120° angle. A passive spring is associated with each arm that enables the robot to be adaptable to different pipe sizes with the adaptability range of 18 cm-56 cm. The arm modules are equipped with actuator modules that move the robot inside pipes and each actuator module includes a gear motor, motor cover, a pair of ball bearings, and a wheel. The passive springs press the wheels on the pipe wall and the gear motors provide traction force to move the robot inside pipes. Fig. 1 shows the CAD design and prototype of the proposed robot. We also addressed the health considerations in design in which the robot is not a source of toxic materials and its components do not react chemically with water during operation. In Table 1, the specifications of the robot are listed. More details on the design and prototyping can be found in [43,44].

**Table 1**. Specification of the Robot.

| Parameter (Unit) | Value | Description |
|---|---|---|
| Application | - | Water Quality Monitoring and Leak Detection |
| Mechanism Type | - | Under-actuated |
| Locomotion Type | - | Wheeled Wall-press |
| Actuator | - | Gear-motor Equipped with Incremental Encoder |
| Prototyping Method | - | 3D Printing |
| Printing Material | - | Acrylonitrile butadiene styrene (ABS) |
| Size-adaptable | - | Yes |
| Operation Environment | - | Water Distribution Systems |
| Size-adaptability Range (in) | 9-20 | Adaptability to Internal Shape of Pipes. |
| Length Range (in) | 7.5-9 | Length of the Robot along Pipe Axis |
| Weight (kg) | 2.23 | - |

There is a notion of resilient machine that is first coined in [45] and formulates the requirements for manufacturing a machine that is reliable and robust against uncertainties and disturbances. We need to consider and take the requirements in our work towards the design of a resilient robot. In this regard, we consider a scenario where the robot gets stuck during operation and fails to mover and investigate the parameters that leads to this failure. The robot is supposed to move in operating distribution systems, and multiple factors affect its motion. In the next section, we present these parameters and analyze the effect of each one. Our goals in simulation the operation condition in pipes are: investigation of the robot interference to flow, the range of drag force applied on the robot, and definition of an extreme condition for the robot to have an estimate for the maximum drag force (Drag force is a resistive force that applied on an object that moves in a fluid). This analysis is useful for long-distance inspection for the robot. We simulate the operation condition with flow simulation and computational fluid dynamics (CFD) work in SolidWorks. The parameters of the flow simulations are based on standards in water distribution systems [15].

## 3. Parameters in Operation Environment of the Robot and Minimum Travel Distance

The opration environment of in-pipe robots is pressurized pipelines where high-velocity flow is present. Hence, we investigate the effect of line pressure and relative velocity of the robot and water flow with CFD work. In our simulations, the robot is located in a 20-in diameter pipe to have the maximum outer diameter, and hence, provides us the maximum or minimum amount of effects of the parameters.

### *3.1. Effect of Line Pressure on the Robot*

In this section, we analyze the effect of line pressure, and to this aim, we performed several flow simulations in different pressures from 100kPa to 500kPa which are the range of line pressure in distribution networks based on standards and the relative velocity of the robot and the water flow is 1.2 m/s. The simulations are shown in Fig. 2, [a]-[e]. The colored arrows show the pressure counter around the robot inside the pipe. The drag forces in simulations are shown in table. 2 and Fig. 3. The results show that there is negligible pressure gradient on both sides of the robot in the pipe (less than 1%), hence, the robot is minimally invasive to water flow during operation that is desirable feature and helps to save power. Moreover, an increase in line pressure decreases the drag force applied to the robot. However, this effect is small; an increase of 400% in line pressure in our simulation, results in a

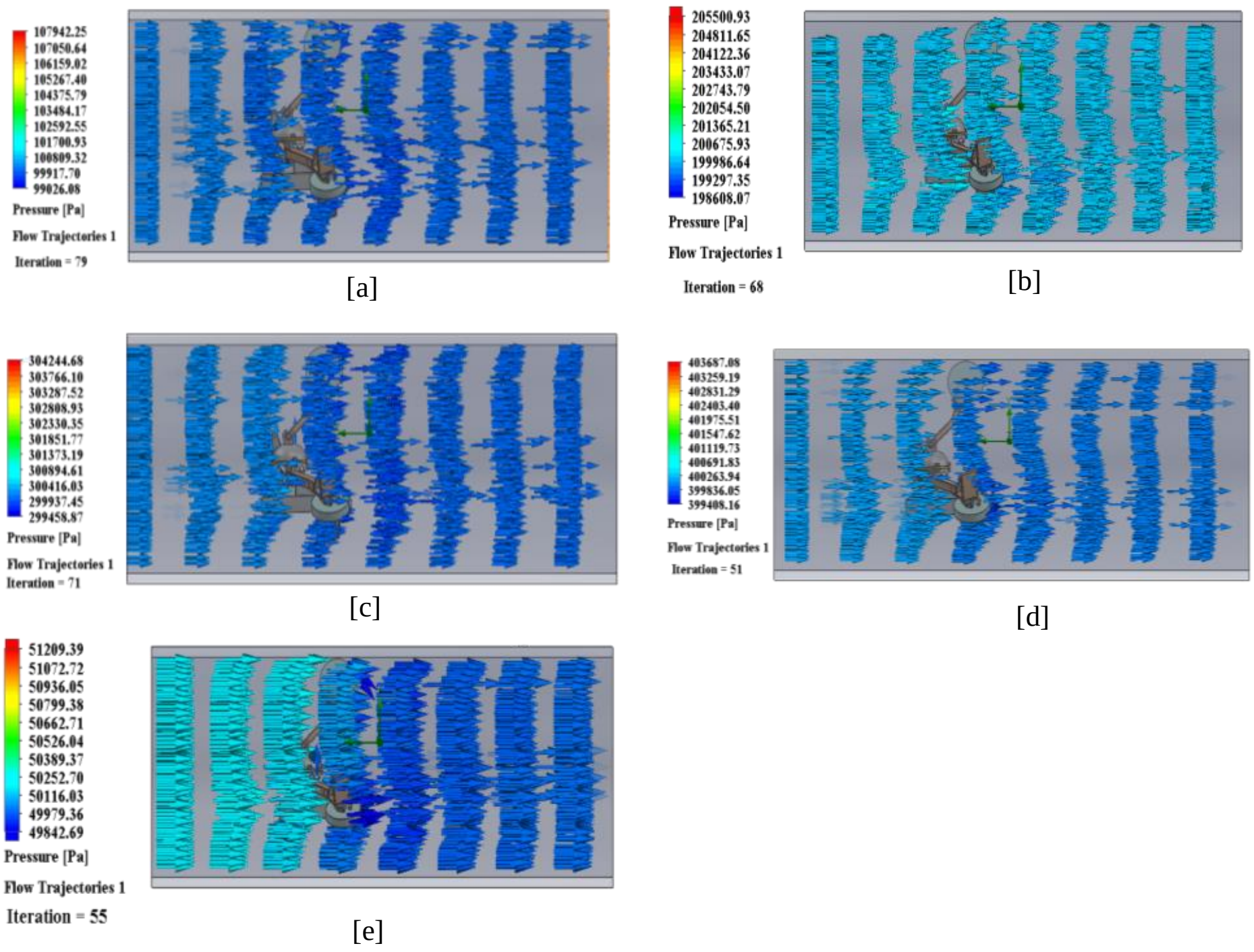

**Fig 2**. Flow simulations to evaluate the effect of line pressure on the drag force. [a] Line Pressure: 100 kPa. [b] Line Pressure: 200 kPa. [c] Line Pressure: 300 kPa. [d] Line Pressure: 400 kPa. [e] Line Pressure: 500 kPa.

27% decrease in drag force that means, there is no meaningful correlation between the line pressure and the drag force. However, the main effect of the line pressure is on the strength of the robot's components and we addressed this in our design in [44].

**Table 2**. Drag Force in Different Line Pressure.

| Line Pressure [kPa] | Pressure Gradient [%] | Drag Force [N] |
|---|---|---|
| 100 | 0.88 | -25.9 |
| 200 | 0.34 | -24.4 |
| 300 | 0.16 | -18.3 |
| 400 | 0.11 | -19.3 |
| 500 | 0.02 | -18.9 |

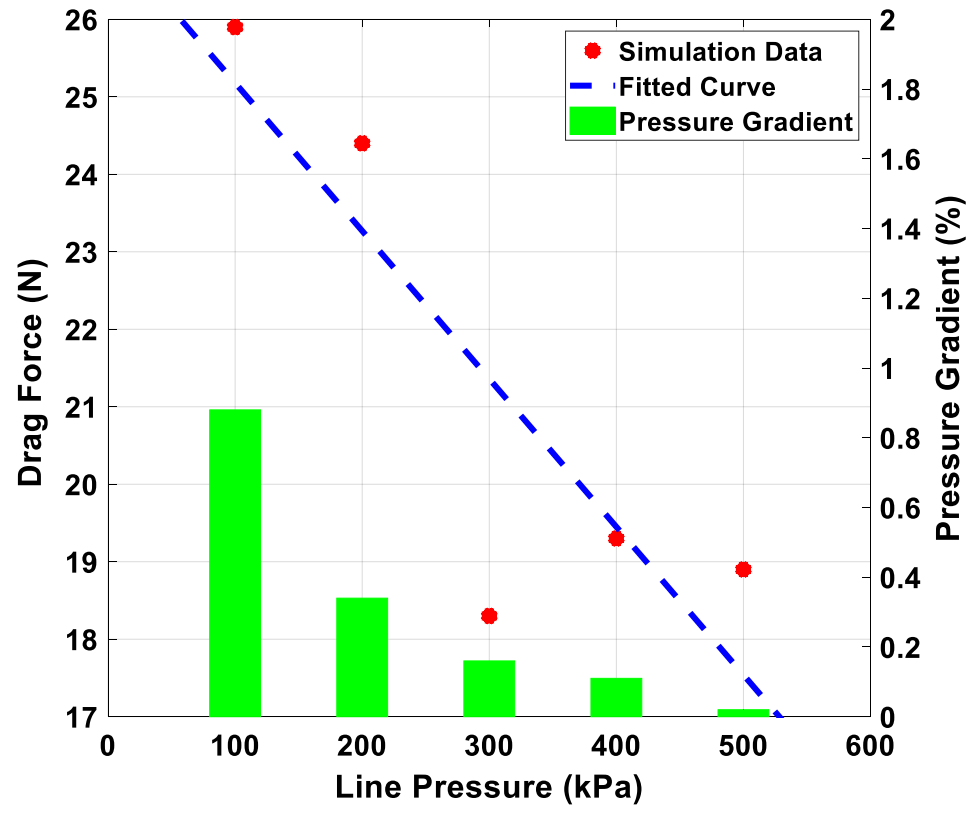

**Fig. 3**. Drag Forces Applied on the Robot in Different Line Pressures

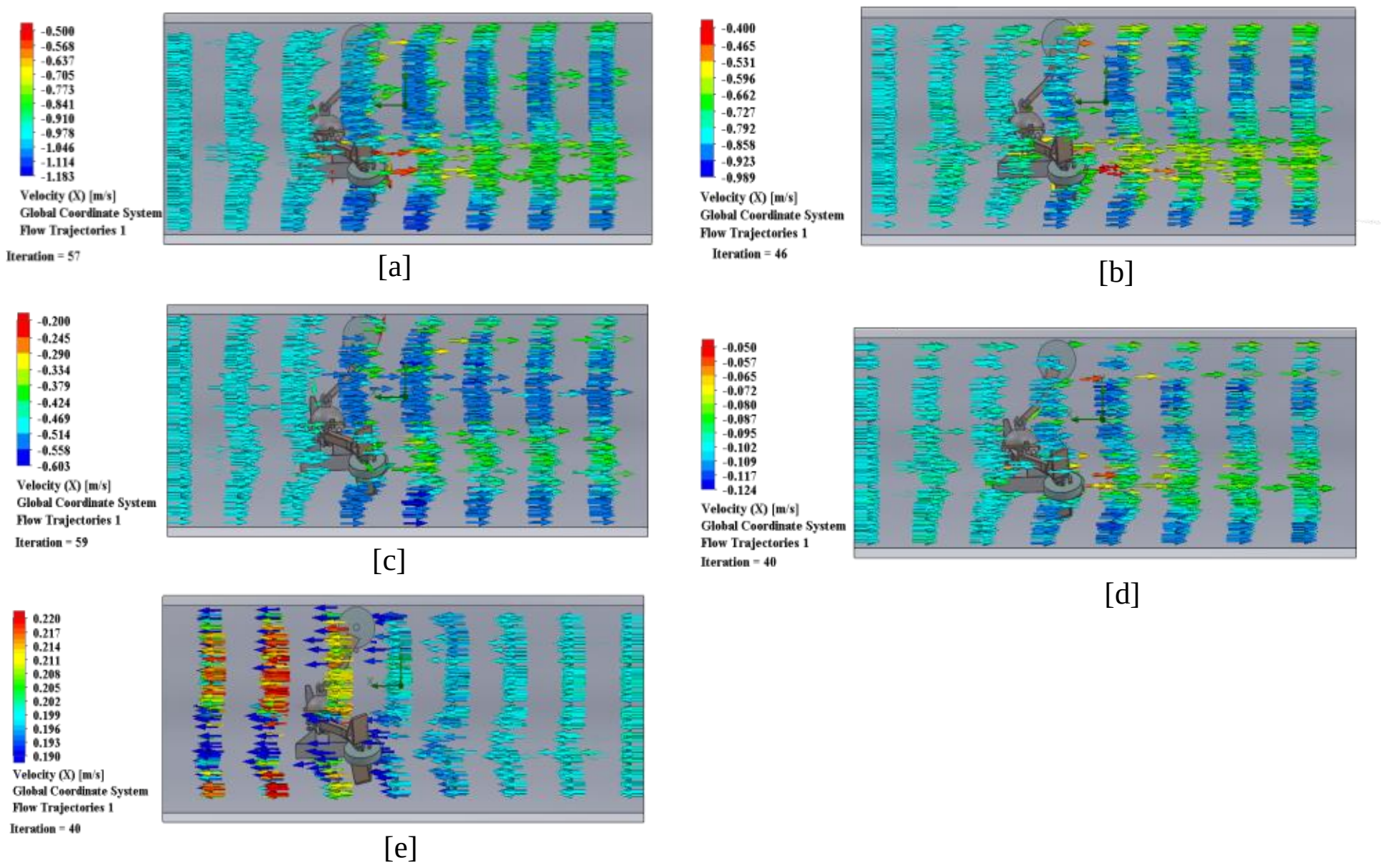

**Fig. 4**. Flow Simulations to Evaluate the Effect of Relative Velocity on the Drag Force with Different Relative Velocities of [a] 1.2 m/s with Oppoosite Directions. [b] 0.8 m/s with Oppoosite Directions. [c] 0.5 m/s with Oppoosite Directions. [d] 0.1 m/s with Oppoosite Directions. [e] 0.2 m/s with Aligned Directions.

### *3.2. Effect of Relative Velocity on Drag Force*

In this part, we evaluate the effect of relative velocity on the drag force. To this aim, we did five iterations of flow simulations with 100 kPa line pressure that had maximum drag force on the robot and different relative velocities. Fig. 4 shows the simulations and the velocity counter around the robot in the pipe. In iterations [a]-[d], the motion directions of the robot and flow are opposite and in iteration [e], they move in the same direction. The velocity differences on both sides of the robot are less than 10%. The other parts are less invasive to the velocity of the flow. So, in terms of velocity, the robot is minimally invasive to the flow that is desirable in terms of power consumption in the robot. The drag force in each iteration of the simulation is listed in Table. 3.

**Table 3.** Drag forces in Different Relative Velocities

| Relative Velocity [m/s] | Flow Velocity Variation on Both Sides of the Robot [%] | Drag Force [N] |
|---|---|---|
| 1.2 | 9.62 | -25.9 |
| 1 | 6.51 | -14.4 |
| 0.5 | 9.21 | -6.2 |
| 0.1 | 7.28 | -0.2 |
| -0.2 (In-direction Motion) | 4.56 | +0.6 |

The results of the simulations are shown in Fig. 5. The results show that a 700% increase in the relative velocity, results in a 5400% increase in drag force applied on the robot that is a huge amount.

### *3.3. Minimum Travel Distance*

Based on the simulation results, the maximum drag force is applied on the robot in the pipelines with 100 kPa and 1.2 m/s relative velocities and the amount of drag force is around 26 N. Since the geometry of the robot is symmetric, each motor is needed to provide around 8.7 N to overcome the drag. It worth mentioning that this is the worst-case scenario for the robot operation and our goal is to find the minimum operation duration. The gear-motor (from Maxon Motors Inc©) at this operating point, draws 0.47 A current and considering the three actuators we have in the robot, the 1.41A. The discharge time of the battery in this current and voltage,12 V, is around 3 hours. So, we have a robot that can move at least 3 hours in the in-service distribution networks in extreme conditions with 0.5 m/s velocities against the flow with 0.7m/s. Hence, it can inspect around 5400 m (i.e. 3.35 mile) of the pipeline with one turn of battery charge with a proper motion control algorithm that is higher than its counterparts [46,47]. In the next section, we propose the motion cocntrol algorith, for the robot.

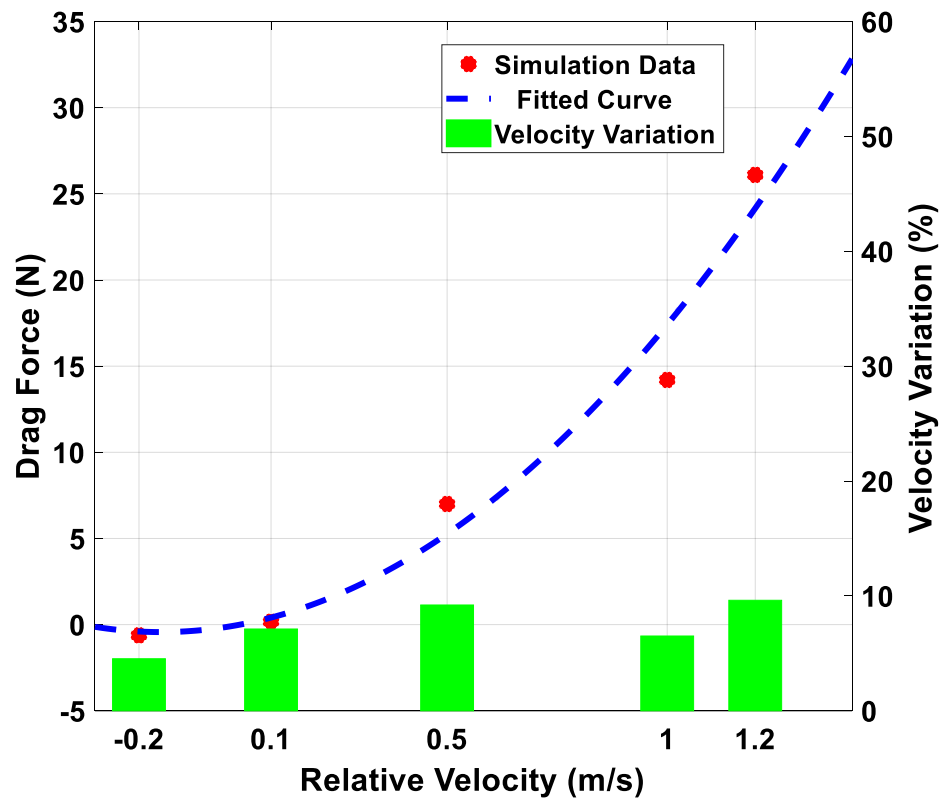

**Fig. 5**. Drag Forces Applied on the Robot in Different Relative Velocities.

## 4. Multi-phase Motion Control Algorithm

Distribution systems comprise pipes with different configurations and sizes. The in-pipe robots needs to negotiate all configurations during operation. Also, it needs to stop at some locations with stabilized configurations to perform sensor measurements. For example, water quality sensors require around one minute time to have a reliable output [48]. To this aim, we design three phases of motion controllers that:

1. In phase 1: Stabilizes the robot with zero velocity at junctions.
2. In phase 2: stabilizes the robot and tracks the desired velocity in the straight path.
3. In phase 3: Steers the robot to the desired directions in non-straight configurations of pipelines (e.g. bends and T-junctions).

The robot switches between three phases of the controllers considering its location in the network and the configuration type (i.e. straight path, junction, and non-straight configuration). In this section, we propose these controllers and validate their functionality and in the next section, we explain how the robot defines its location in the network and the configuration type.

### *4.1. Phase 1: Stabilizer Controller*

The robot needs to stop at junctions with stabilized configuration to perform a specific task in the pipe (e.g. water quality monitoring). We define $x_s = [\phi \quad \dot{\phi} \quad \psi \quad \dot{\psi}]^T$ as stabilizing states (Fig. A.1) and the task in the stabilization is to keep the stabilizing states near zero during operation. As for the stabilizer controller, we use linear quadratic regulator (LQR) controller which is a state feedback controller and shows promising results in stabilization [49]. However, to design this controller, state space representation of the system is needed and the dynamical equations of the system are highly nonlinear (Eqs. (A3)-(A5) in appendix A).

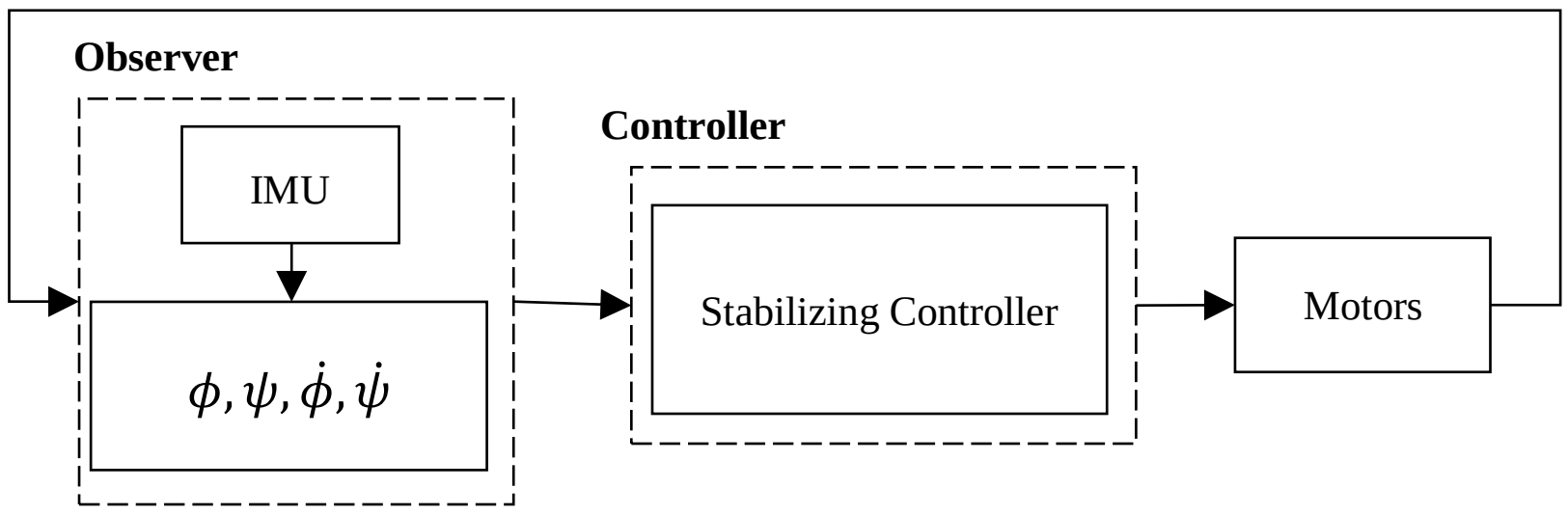

**Fig. 6**. Stabilizer Controller in Junctions.

Hence, we linearize the dynamical equations of the robot around the equilibrium point and construct the system's auxiliary matrices and the gain matrix of the LQR controller is computed (Full calculations are presented in appendix B). The observer in the proposed controller includes an inertial measurement unit (IMU) which comprises an accelerometer and gyroscope. The output of the IMU sensor is noisy and to have reliable outputs from the IMU, we used Mahony filter [50] to calculate the orientation of the robot ($\phi$ and $\psi$). Mahony filter computes the orientation of the robot by fusing the data from the accelerometer and gyroscope of the IMU. Also, $\dot{\phi}$ and $\dot{\psi}$ are provided directly from the gyroscope. The controller in this phase is shown in Fig. 6. Also, the performance of the controller in this phase is evaluated with experimental results (four iterations) in Fig. 7. In this experiment, An BMI160 6DOF 6-axis is located in the central processor and the robot is located in a 14-in diameter pipe. Four initial deviations for $\phi$ and $\psi$ are given to the robot (see table 4) and the controller is supposed to cancel them. The results show that the initial deviations are cancelled within one second since the motion is started and the stabilizing states are kept at zero for the rest of motion. Hence, the controller is able to cancel the deviations within one seconds.

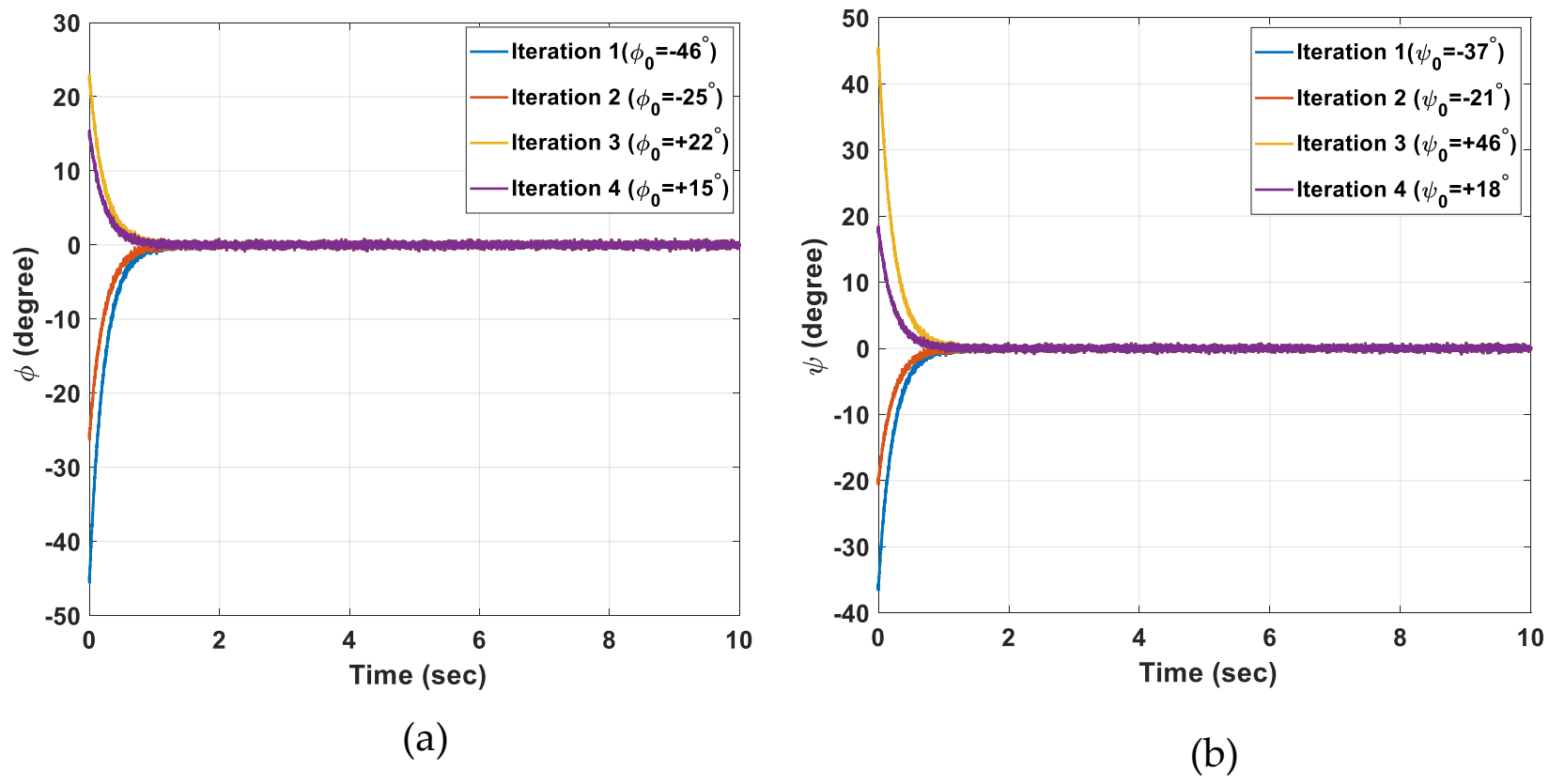

**Fig. 7**. Performance of the Stabilizer Controller in a 14-in Diameter Pipe. (a) $\phi$ (degree). (b) $\psi$ (degree).

**Table 4.** Specifications of the Experiments for the Stabilizer Controller.

| Iteration | Initial Value for $\phi$ ($\phi_0$) (degree) | Initial Value for $\psi$ ($\psi_0$) (degree) |
|---|---|---|
| 1 | -46 | -37 |
| 2 | -25 | -21 |
| 3 | +22 | +46 |
| 4 | +15 | +18 |

### 4.2. *Phase 2: Stabilizer-velocity Tracking Controller*

In the straight paths, the robot needs to track a desired linear velocity along pipe axis in addition to the stabilization. Hence, we design a velocity controller and add it to the stabilizer controller that developed in phase 1. The linear velocity of the robot is directly computed from the angular velocity of the wheels, if the wheels have approximately equal velocity. The relation can be written as:

$$v = R\omega \tag{1}$$

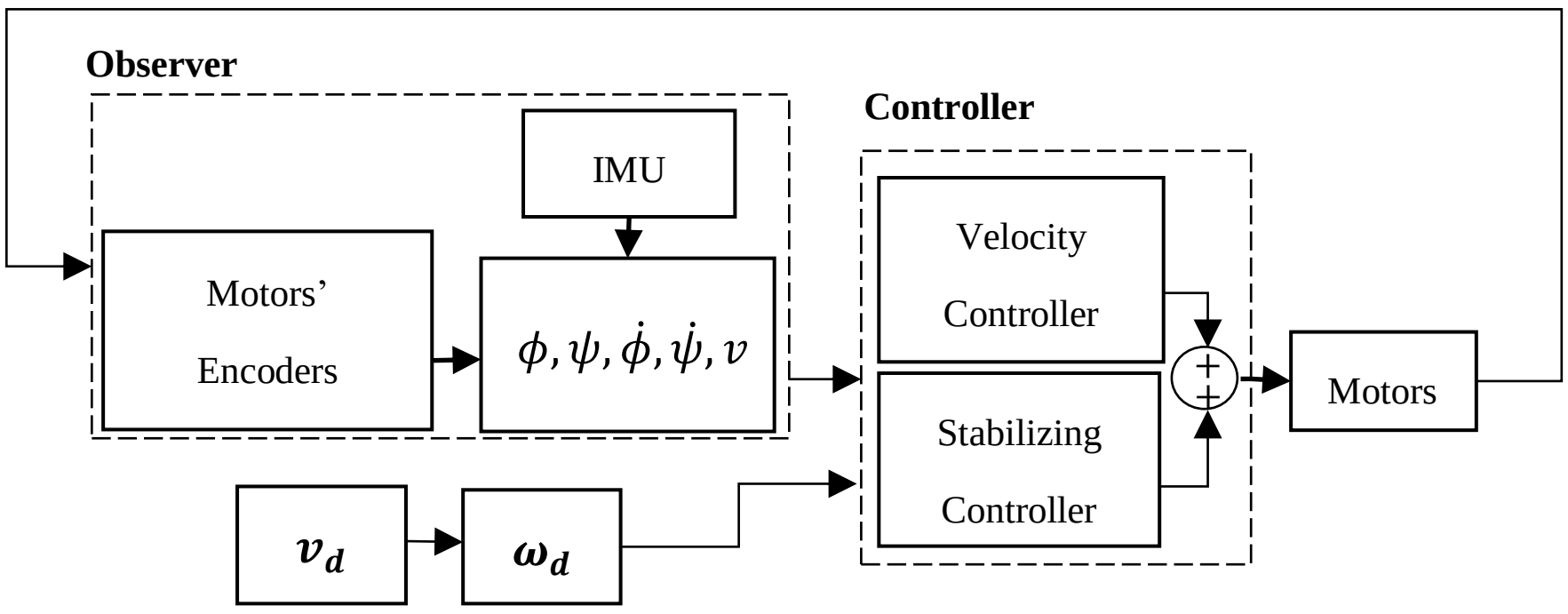

**Fig. 8**. Phase 2 of the controller to stabilize the robot and track the desired velocity in straight paths.

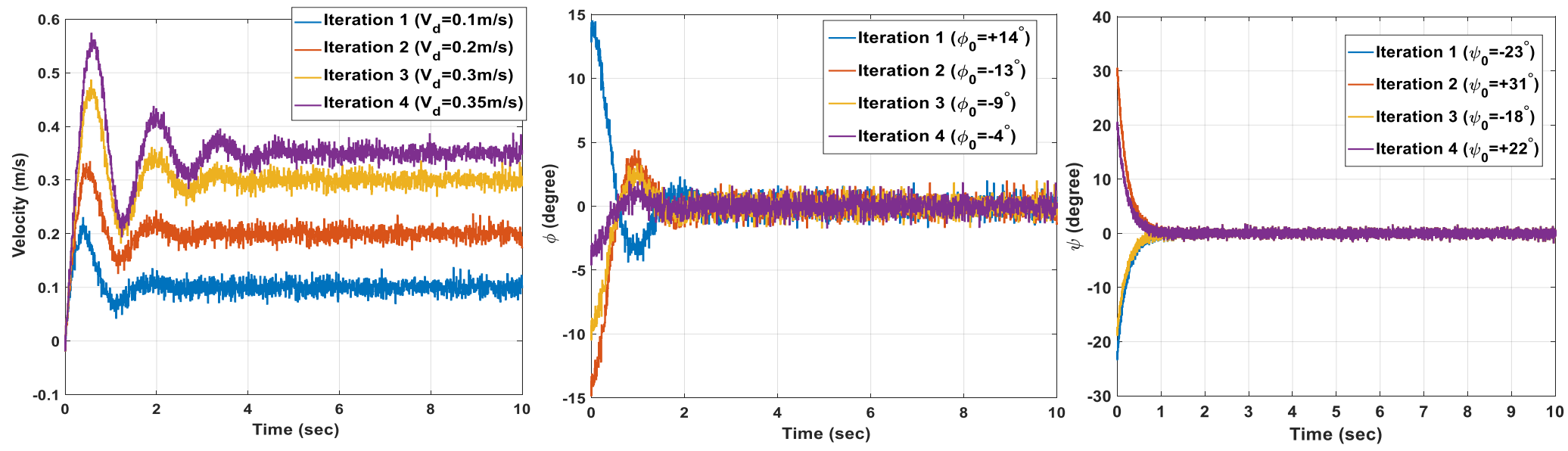

**Fig. 9**. Performance of the Stabilizer-velocity Tracking Controller in a 14-in Diameter Pipe. (a) Linear velocity (m/s). (b) $\phi$ (degree). (c) $\psi$(degree).

where $v$ is the linear velocity of the robot, $R$ is the radius of the wheel, and $\omega$ is the angular velocity of the wheel. Hence, desired angular velocities of the wheels, $\omega_d$ can be computed with desired linear velocity, $v_d$. The motors' encoders measure the angular velocity of the wheels and three proportional-integral-derivative (PID) controllers, controls the velocity of the wheels. In appendix B, the parameters of the PID controllers are shown. The controller is shown in Fig. 8. The observer in this controller includes the IMU and three motor encoders that connected at the end of the gear motors. We also evaluated the performance of the controller with experiments in a 14-in diameter pipe. The experiments are repeated in four iterations and in each iteration, an initial deviation for stabilizing states and desired linear velocity are considered (Table 5). The experiments show that the initial deviations are canceled within two seconds since the motion is started and the robot reaches the desired velocity within four seconds of motion start. Hence the proposed controller in a straight path meets the requirements of motion in a straight path (see Fig. 9) [43].

**Table 5.** Specifications of the Experiments for the Stabilize-velocity Trackingr Controller.

| Iteration | Desired Velocity [m/s] | Initial Value for $\phi$ ($\phi_0$) [degree] | Initial Value for $\psi$ ($\psi_0$) [degree] |
|---|---|---|---|
| 1 | 0.10 | +14 | -23 |
| 2 | 0.20 | -13 | +31 |
| 3 | 0.30 | -9 | -18 |
| 4 | 0.35 | -4 | +22 |

### *4.3. Differential Motion Controller*

The robot needs to steer to the desired direction at non-straight configurations of pipelines. To this aim, we design a controller that facilitates the desired amount of rotation at junctions with differential drive motion [51]. The trajectory generates differential motion for the robot by assigning different desired angular velocities for the wheels ($\omega_{1d}, \omega_{2d},$ and $\omega_{3d}$). Fig. 10 shows the coordinate system attached on the robot, the wheels numbers, and the definition of the differential motion type. Also, table 6 shows the desired velocities for the wheels in each differential motion type. The maximum and minimum velocities stand for unequal velocities (where maximum velocity is greater than minimum velocity) and their value is defined based on the desired linear speed at passing the non-straight configuration. The controller is shown in Fig. 11 that includes a trajectory generator, an observer, an error-check submodule, and velocity controllers for the wheels. The observer in the developed controller (see Fig. 11) provides information about the orientation and angular velocities of the wheels ($\omega_1, \omega_2, \omega_3$), and the error-check sub-module monitors the status of rotation (of the robot) in the non-straight configuration during rotation. To evaluate the performance of the robot and the controller in this phase, we use ADAMS/MATLAB co-simulation [51]. ADAMS is dynamic simulation software and can model the parameters of the robot and distribution network. Hence, the robot and the pipe are simulated in ADAMS and the controller is simulated in MATLAB Simulink Toolbox. The controller variables are updated in MATLAB and system measurements are updated in ADAMS in real-time. To this aim, the CAD design of the robot and the pipe that are designed in the SolidWorks are imported in ADAMS and all variables of the robot (i.e. joints, forces, gear motors) are defined. Then the control inputs (motor voltages) and the system outputs (rotation around x, y, and z axes, and angular velocities of the wheels) are defined. A system plant is then generated with this architecture (system's inputs and outputs) and connected to the MATLAB Simulink and a control model is constructed based on the developed control algorithm and the system plant.

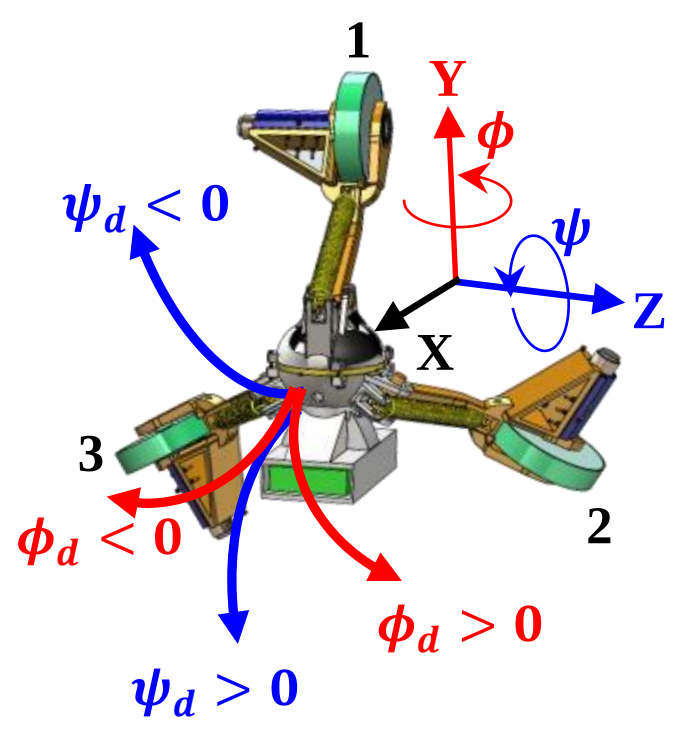

X-axis: Pipe Axis

**Fig. 10**. Coordinate System on the Robot, Wheel Numbers, and the Definition of Differential Motions.

**Table 6.** Different Types of Deffiential Motion and the Requirements for the Wheels Motion in each Type.

| Differential Motion Type | Desired Velocity for Wheel 1 ($\omega_{1d}$) | Desired Velocity for Wheel 2 ($\omega_{2d}$) | Desired Velocity for Wheel 3 ($\omega_{3d}$) |
|---|---|---|---|
| $\phi_d > 0$ | 0.5 (Maximum Velocity + Minimum Velocity) | Minimum Velocity | Maximum Velocity |
| $\phi_d < 0$ | 0.5 (Maximum Velocity + Minimum Velocity) | Maximum Velocity | Minimum Velocity |
| $\psi_d > 0$ | Minimum Velocity | Maximum Velocity | Maximum Velocity |
| $\psi_d < 0$ | Maximum Velocity | Minimum Velocity | Minimum Velocity |

We evaluate the performance of the controller in phase 3 in bends and T-junctions in which are common non-straight configurations in pipelines [52].

### *4.3.1. Controller Performance at 90-degree Bend*

In this part, the performance of the robot and the controller in bends is shown. The robot is supposed to track the desired direction in Fig. 12a in which it needs to rotate 90° clockwise around the y-axis ($\phi_d = -90°$). The inner radius of the bend, $R_i$, is 12-in and the outer diameter, $R_o$, is 24-in. The desired linear speed is 10 in/s and the desired angular velocities for the wheels are: $\omega_{1d} = 34.5$ rpm, $\omega_{2d} = 46$ rpm, and $\omega_{3d} = 23$ rpm. The co-simulation is performed and the amount of rotation around axes at the end of the motion are shown in Fig. 12b. The amount of rotation at the end of the motion shows complete rotation around the desired axis (y-axis) with small rotation around the x-axis ($\approx -0.66°$) and z-axis ($\approx -13.33°$). The rotation around x-axis is almost zero as expected. However, the rotation around z-axis is not zero, since after the robot rotated completely, it switches to a stabilizer-velocity tracking controller, these small deviations are canceled as shown Fig. 9. We repeated our simulations and validated that the robot and the controller can cover the bends with diameters ranging from 9-in to 22-in.

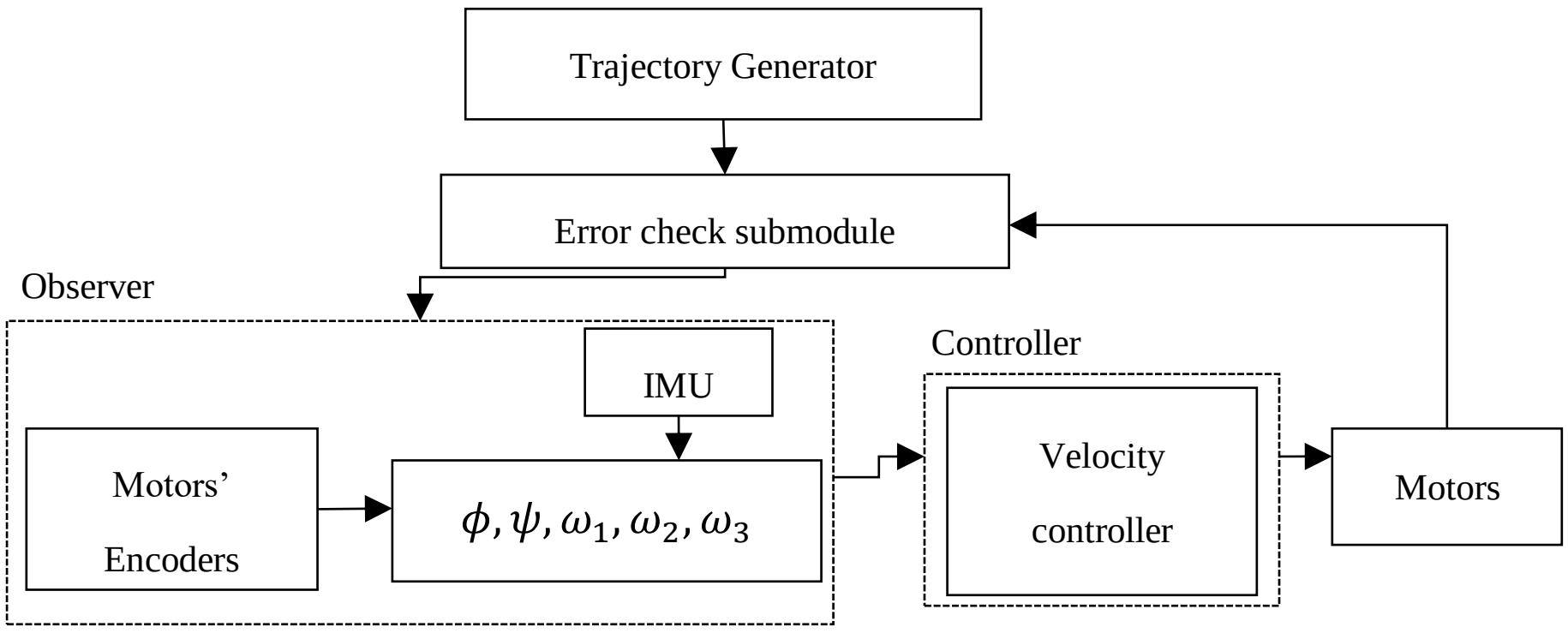

**Fig. 11.** The proposed controller that enables the robot to have desired amount of rotation around axes.

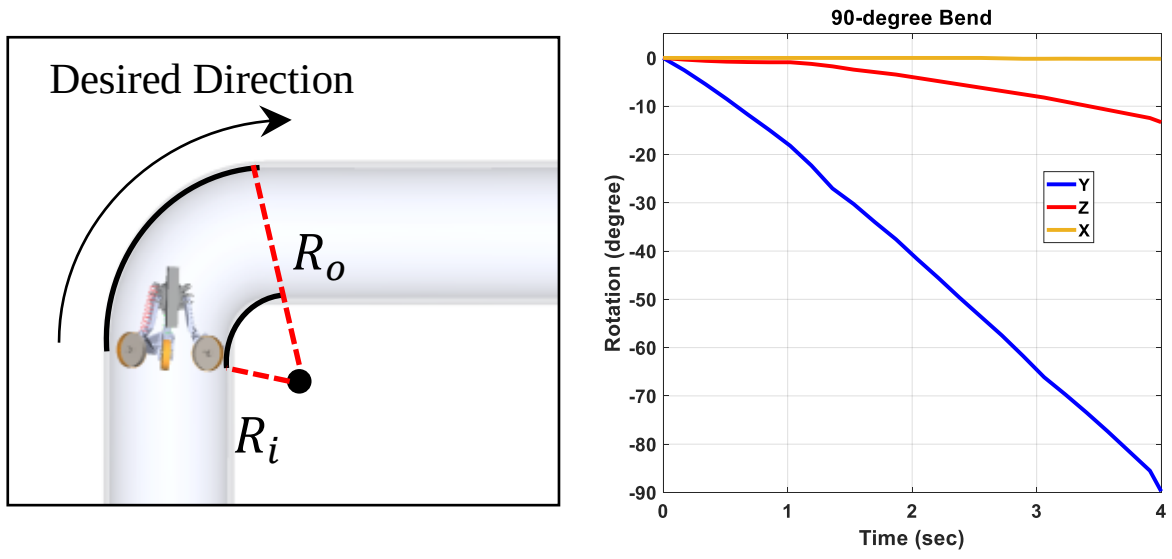

[a] [b]

**Fig. 12.** [a] Desired Direction in 12-in Diameter Bend ($R_i$=12-in, $R_o$=24-in). The robot needs to rotate 90° clockwise around y-axis ($\phi_d = -90$). [b] Amount of rotation around axes during rotation. $\omega_{1d} = 34.5$ rpm, $\omega_{2d} = 46$ rpm, and $\omega_{3d} = 23$ rpm.

### *4.3.2. Controller Performance at T-junction*

The robot is supposed to track the desired direction shown in figure 13a. In this scenario, the amount of rotation is 90° counter-clockwise ($\phi_d = +90°$). $R_i$, is 14-in and $R_o$ is 28-in. The desired linear speed is around 15 in/s and the desired angular velocities for the wheels are: $\omega_{1d} = 73$ rpm, $\omega_{2d} = 49$ rpm, and $\omega_{3d} = 97$ rpm. The amount of rotation at the end motion around each axis shows complete rotation around the desired axis with small rotation around the x-axis ($\approx -1.21°$) and z-axis ($\approx +22.46°$) (see Fig. 13b) that are canceled when the robot switches to stabilizer-velocity tracking controller. We repeated our simulations and derived the T-junction coverage to be 9-in and around 15-in. In our simulations, we found that the inner radius shape of the T-junction plays an important role toward reliable motion. The more stiff the inner shape of the T-junction, the less reliable motion. However, the pretension in the springs make the arms to stick to the pipe wall, but it leads to rotation around x-axis that is undesirable and the coordinate system needs to be updated. The rotations around the z-axis can be cancelled when the robot switched to the stabilizer-velocity tracking phase.

In this section, we designed a multi-phase motion control for our proposed robot that enables maneuverability with stabilized motion in complicated configurations of pipeline that is has been a challenge for in-pipe robots [39,46,53–60]. However, switching between different phases of motion controller remains a challenge to solve. In other words, the robot needs to localize itself in the network to choose the appropriate phase of motion controller. In the next section, we propose a method that enables the robot to localize itself during operation and navigate through complicated configurations of pipelines.

## 5. Smart Navigation Using Particle Filtering

### *5.1. Particle Filtering Method*

Particle filtering (PF) is an estimation method for the parameter(s) of a system that we cannot measure directly by sensors. The PF performs estimation using the relationship between measurable parameters(s) and unknown parameters. In this method, initially, a random sample (particles) with arbitrary weights are considered for the unknown parameter(s). In the next time step, the weight of the particles are updated based on the new measurement from the environment. Next, the particles are resampled based on the updated weights. At final step, the resampled particles are propagated in time. The process is repeated until the particles converge to a value that is the estimation of that unknown parameter. Kalman filter estimates the unknown parameters like PF, however, in the Kalman filter, distribution of the measured parameter(s) and the estimated distribution of the unknown parameter(s) should be Gaussian [61] while the PF can estimate the parameters with non-Gaussian distributions as well. The capability of PF in estimation of parameters with non-Gaussian distributions makes it a useful tool for localization in in-pipe robots. In this method, a sensor that provides the information from the surrounding environment of the robot is used (e.g. ultrasonic sensor). Algorithm 1 presents the PF method we use for localization of the robot. To this aim, we uniformly randomize both the location and of the robot across the entire operation map $< x_{t-1}^{j}, w_{t-1}^{j} >$. It worth mentioning that higher number of random points leads to higher accuracy at the expense of lower convergence speed. The weight of the particles are updated using the information from the ultrasonic sensor $(z_t)$ and the particles are resampled. Next, random noise is added to the robot motion direction (system input) and apply the input $(u_t)$ to the resampled particles (with updated weights). $\{S_{t-1} = < x_{t-1}^{j}, w_{t-1}^{j} >, \ u_t, \ z_t\}$ is the set that includes $< x_{t-1}^{j}, w_{t-1}^{j} >$ that is the discrete distribution of locations $x$ with weights $w$ , $u$ is the system input, and $z$, new measurement of the ultrasonic sensor. All parameters are in time step $t-1$. It should be mentioned that the robot needs the map of the operation.

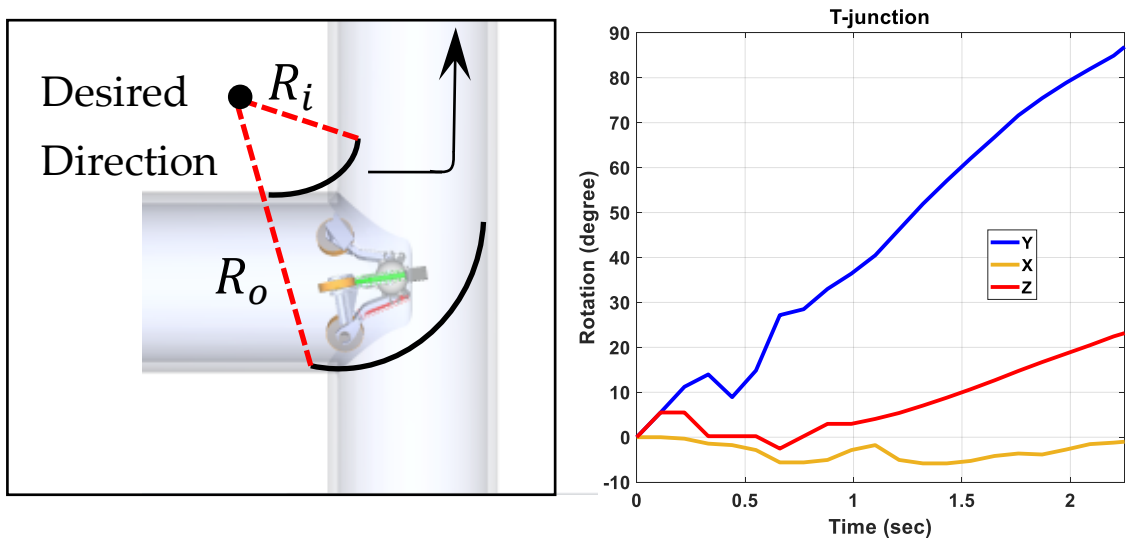

[a] [b]

**Fig 13.** [a] Desired Direction in 14-in diameter T-junction ($R_i$=14-in, $R_o$=28-in). The robot needs to rotate 90° counter- clockwise around y-axis ($\phi_d = +90$) [b] Angular Velocities of the robot around axes and also linear speed (dashed line) during rotation. $\omega_{1d} = 73$ rpm, $\omega_{2d} = 49$ rpm, and $\omega_{3d} = 97$ rpm.

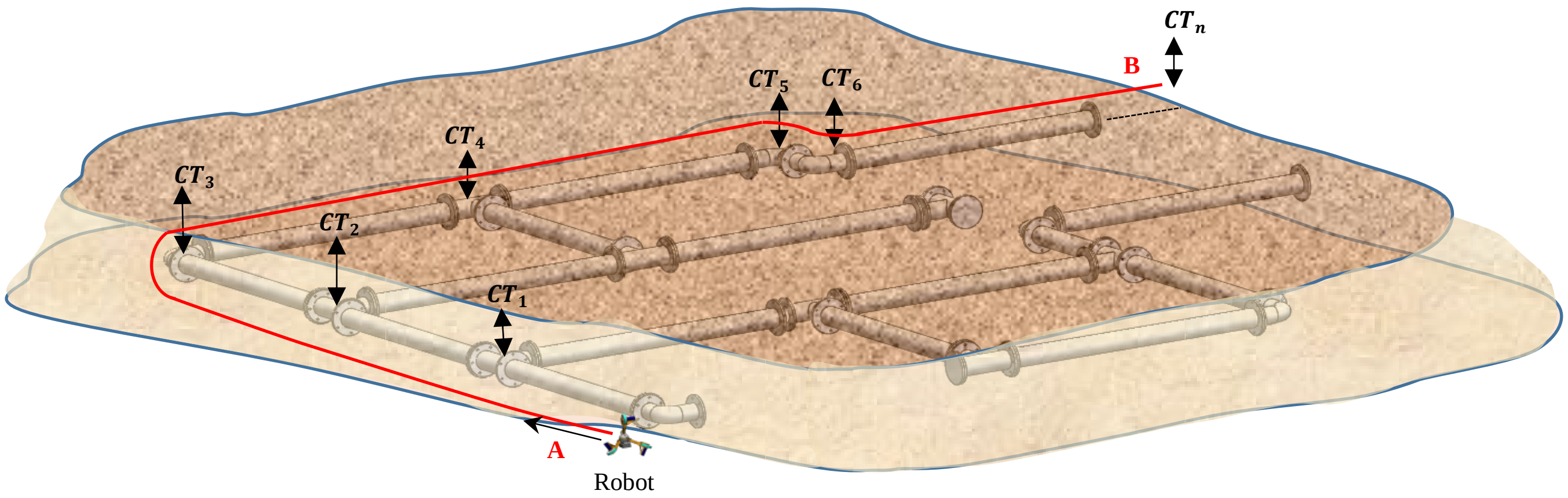

**Fig. 14.** The proposed robotic network with the array of non-straight configurations of pipeline that the robot needs to pass through. This array acts as the map of the operation and the robot localize itself with the given map, ultrasonic sensor, and oarticle filtering method. In this figure, the robot moves from point **A** to point **B**.

| **Algorithm 1: Particle Filtering Method** | | |
|---|---|---|
| **1** | $S_t = \emptyset, \eta = 0$ | |
| **2** | **for** $i$ from 1 to $n$ **do** | |
| **3** | Sample index $j(i)$ from the discrete distribution given by $w_{t-1}$ | |
| **4** | Sample $x_t^i$ from $p(x_t \mid x_{t-1}, \boldsymbol{u_t})$ using $x_{t-1}^{j(i)}$ and $\boldsymbol{u_t}$ | |
| **5** | $w_t^i \leftarrow p(z_t \mid x_t^i)$ | *Recompute the weights by likelihood* |
| **6** | $\eta \leftarrow \eta + w_t^i$ | *Update normalization factor* |
| **7** | $S_t \leftarrow S_t \cup \{< x_t^i, w_t^i >\}$ | *Add the updated particle to the set* |
| **8** | **end for** | |
| **9** | **for** $i$ from 1 to $n$ **do** | |
| **10** | $w_t^i \leftarrow w_t^i / \eta$ | |
| **11** | **end for** | |

### *5.2. Proposed Smart Navigation Method Using Particle Filtering and Multi-phase Controller*

We implement the PF for robot localization and navigation. As mentioned before, to use the particle filtering method, the robot needs the map of operation which means the path the robot needs to travel to reach from point **A** to point **B** (Fig. 14) [62]. We provide the map by dividing the pipeline configurations (that the robot needs to pass through) to straight and non-straight paths (Fig. 14). Then an array is constructed that includes the non-straight configuration types of the operation. The array is represented as:

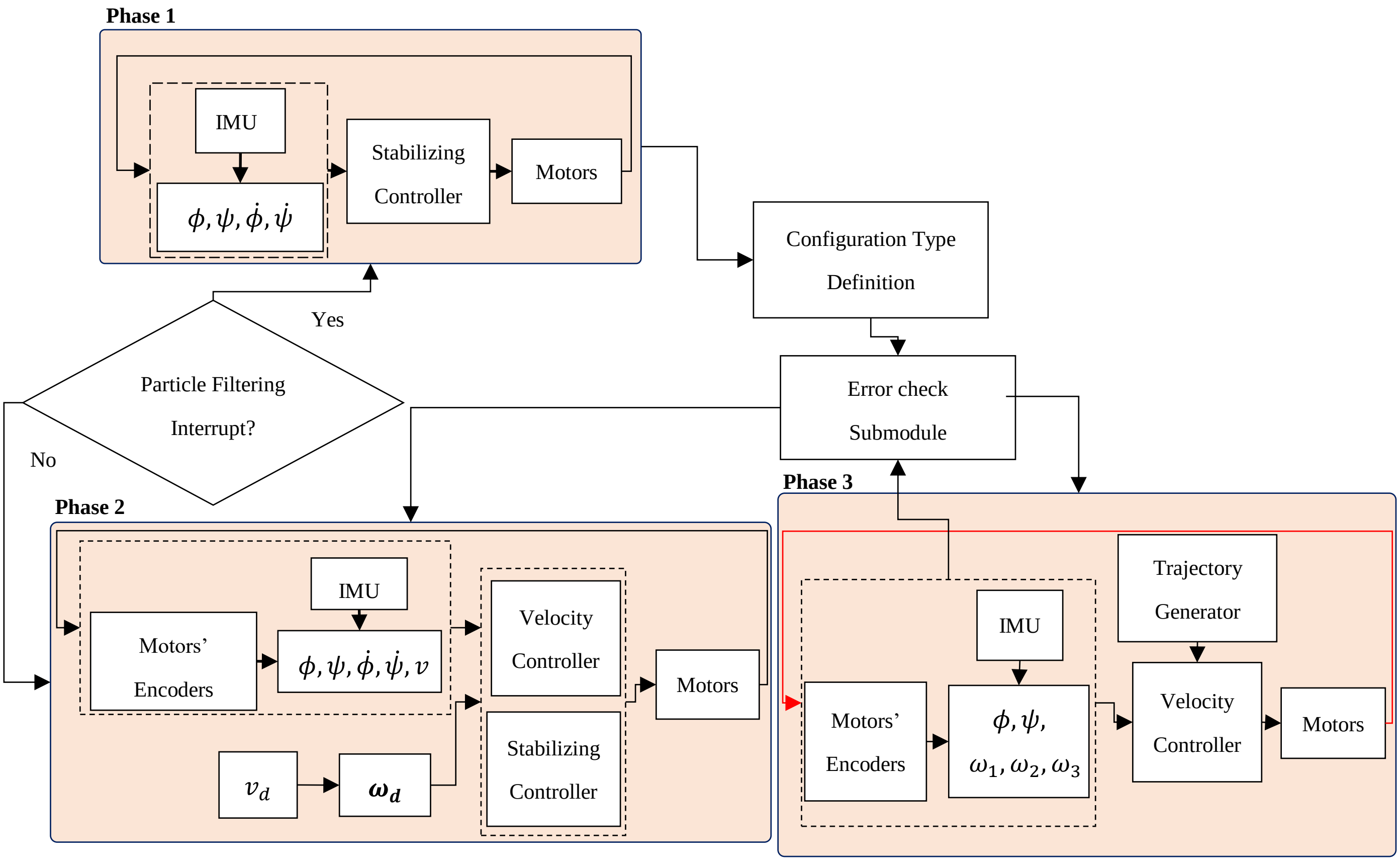

**Fig. 15**. Synchronized Multi-phase Motion Control Algorithm with Particle Filtering Localizer.

$$\mathbf{CT}=[\boldsymbol{CT_1} \quad \boldsymbol{CT_2} \quad \ldots \quad \boldsymbol{CT_n}] \tag{2}$$

In Eq. (2) $\boldsymbol{CT_i}$ defines the configuration type of $\boldsymbol{i^{th}}$ non-straight configuration and the desired new path. For example, $\boldsymbol{CT_1}$ is a T-junction and the desired path for the robot in this configuration is straight. We use an ultra-sonic sensor that measures distance from the front obstacle(s) to have information about the robot's surrounding environment (i.e. $z_i$ in previous section). In terms of navigation, we defined three phases of motion control for the robot in Section 4. In a straight path, the robot moves with phase 2, in non-straight configuration, the robot first stops with stabilized configuration (phase 1) and then steers to the desired direction with phase 3. Switching between three phases is facilitated by the proposed PF method. Fig. 15 shows the synchronized three-phase motion controller and the particle filtering method.

### *5.3. Experiment*

To evaluate the performance of the proposed method, we experimented in which the robot is placed in a 14-in diameter PVC pipe and an HC-SR04 ultrasonic sensor module is attached in front of it that measures the distance from the front obstacle (Fig. 16a). The MCU is Arduino Mega 2560 and the motor drivers are L298N. Our goal in this experiment is to find the capability of switching between stabilizer-velocity tracking controller and stabilizer controller. The robot moves with phase 2 of the controller and when an interrupt from the particle filtering occurs, it switches to phase 2 that is the stabilizer controller. In this experiment, the interrupt is supposed to occur in 14-in the distance from the end of the pipe which indicates the presence of junction (Fig. 16b). The robot moves with phase 2 of the controller in a straight path with 10 cm/s linear velocity (see Fig. 16c). When it reached to the 14-in the distance of the end of the pipe, stopped with stabilized configuration. Hence it has switched from phase 1 to phase 2 of the motion controller. The experiment proves that the reliability of the proposed localizing and navigating method for the in-pipe robots. We repeated the experiments in different desired linear velocities (20 cm/s, 30 cm/s) and found out that higher linear velocities decreases the accuracy of the location where the robot is supposed to stop. This is because the particle filter and ultra-sonic sensor has delay to specify the junction. However, the accuracy reduction is less than 20% that can be improved with other ultra-sonic sensors with faster response time.

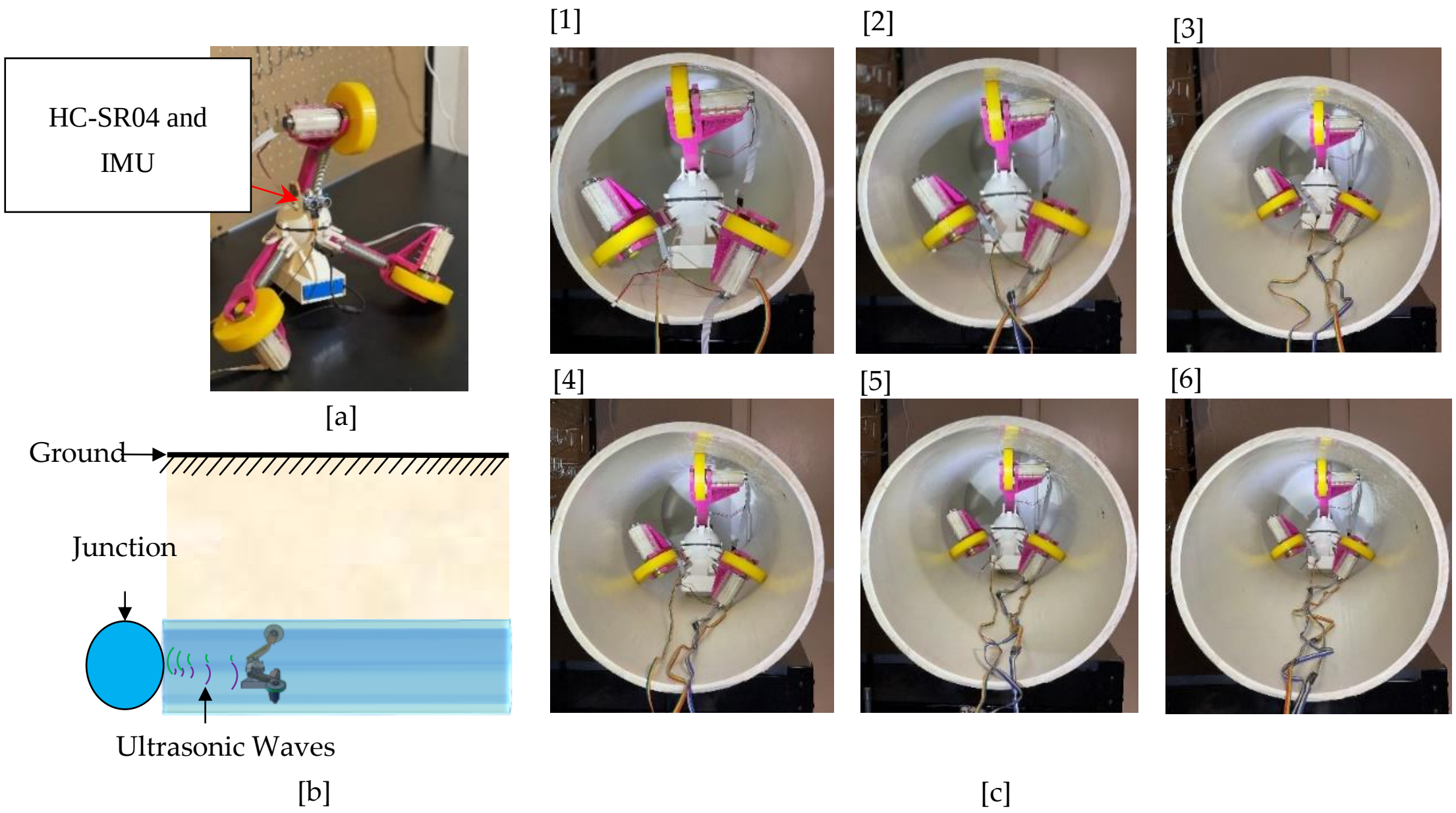

**Fig. 16**. [a] Prototype of the Robot Equipped with HC-SR04 Ultrasonic Sensor. [b] Schematic of the Robot in Pipe and Junction Location. [c] Sequences of motion of the robot in 14 in PVC pipe. The robot starts with the stabilizer-velocity tracking controller and switches to the stabilizer controller when its distance from the front obstacle is around 14-in.

## 6. Discussion and Conclusion

The proposed robot is modular, lightweight (i.e. 2.23 kg), and maneuverable with one set of actuator module. The robot is agile and can move with rather high velocities with gear motors and direct contact power transmission mechanism. In the central processor, a space is considered for the sensor module that enables the robot to be used for multiple tasks like water quality monitoring, leak detection, and condition assessment in water distribution networks. The under-actuated mechanism along with three phases of motion control enables the robot to move in complicated configurations of in-service networks with its own power supply. Operation conditions of the robot are simulated and the effect of line pressure and relative velocity are defined. Considering these parameters, we defined an extreme condition where the drag force is maximum on the robot. Based on the drag force, the minimum inspection distance is defined that is around 5400 m of pipeline with one turn of battery charge with the condition the robot moves in a 20-in diameter pipe with 50 cm/s in opposite direction of the flow with 70 cm/s velocity. The proposed navigation method based on particle filtering and the ultrasonic sensor enables the robot to localize and navigate through the straight and non-straight configurations of pipelines. In this method, the robot is programmed with the map of operation that is an array of non-straight configurations that the robot needs to path through during operation. With the ultrasonic rangefinder, the map, and the particle filter, it can localize itself in the network and also, define the configuration type. In addition, we combine the localizer method with the three phase motion control algorithm to enable the robot to navigate through the pipelines. In this method, the robot moves with the stabilizer-velocity tracking controller in a straight path and when the localizer method defined a non-straight configuration, it stops there with stabilized configuration to perform sensor measurement. After sensor measurements are completed, it steers to the desired direction. When the robot rotated completely, it is again in the straight path and the process continues until the robot reaches to the desired extraction point.

The proposed self-powered robot along with the localization and navigation method, enables long distance inspection in complicated configurations of pipelines with smart navigation that has been a challenge in this field.

We linearized the non-linear equations of the robot around an equilibrium point to design the stabilizer controller that means the stabilizer controller can negotiate small obstacles that are near the equilibrium point. However, there may be some large obstacle in the pipelines that the stabilizer controller is not able to negotiate or lock one wheel from rotation. In our future work, we plan to design a self-rescue mechanism that composes of three auxiliary motors that are located on the central processor and retract the arms. Also, we plan to design an obstacle-avoidance controller for the self-rescue mechanism.

**Conflict of Interest Statement**

The authors declare no conflict of interest during this research.

**Acknowledgements**

This research is suported by Texas A&M Engineering Experiment Station (TEES) internal financial source by Dr. M Katheirne Banks.

## Appendix A.

Fig. A.1 shows the free body diagram of the robot inside pipeline. $F_i,\ i = 1{,}2{,}3$ are traction forces that the motors generate. We have:

$$F_i = \frac{\tau_i}{R}\ \ i = 1{,}2{,}3 \tag{A.1}$$

In Eq. (A.1), $\tau_i$, $R$, and $i$ are motor torque, wheel radius and the wheels numbers, respectiveluy. The motor torque ($\tau$) is liearly computed with current pass through it. We can express the governing equation of the gear motos with the following set of equations:

$$\begin{gathered}\frac{v_{co}}{L_m} - \frac{v_e}{L_m} - \frac{R_m}{L_m} i_m = \frac{di_m}{dt} \\ v_e = K_v \dot{\vartheta} \\ \tau_m = K_v i_m \\ \frac{n^2}{I_l + n^2 I_R} \tau_m = \ddot{\vartheta}\end{gathered} \tag{A.2}$$

The parameters in Eq. (A.2) are described in table A.1.

**Table A.1**

Parameters in Gear-motors Dynamical Equations

| Parameter [unit] | Description |
|---|---|
| $v_{co}$ [V] | Applied Voltage on the Motor |
| $v_e$ [V] | Back EMF Voltage |

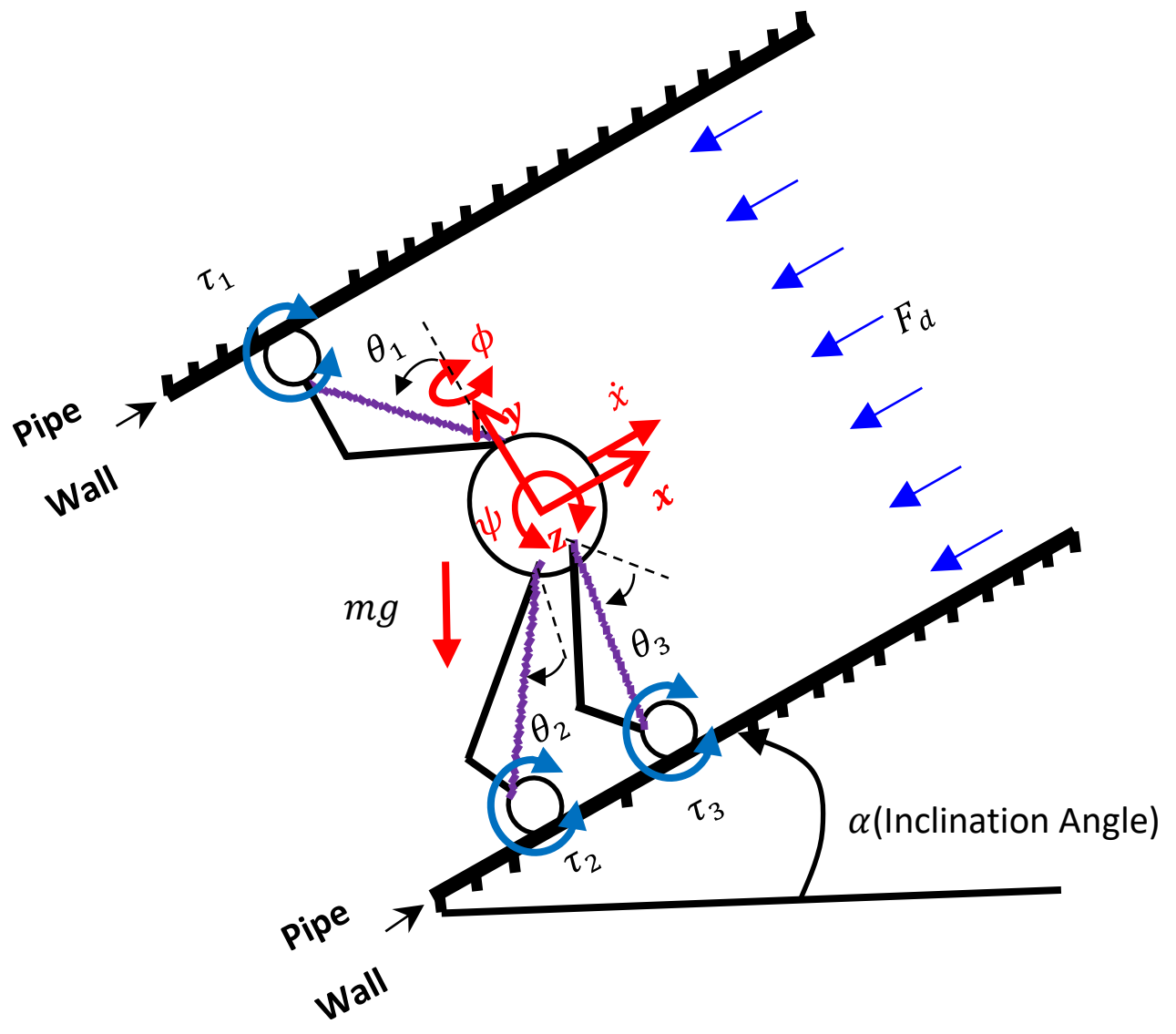

**Fig. A.1**. Free Body Diagram of the Robot in a Pipe with an Inclination Angle, $\alpha$.

| | |
|---|---|
| $i_m$ [A] | Current Pass Through Motor |
| $K_v$ $[min^{-1}.V^{-1}]$ | Speed Constant |
| $L_m[H]$ | Terminal Inductance |
| $R_m[\Omega]$ | Terminal Resistance |
| $\dot{\vartheta}$[rad/s] | Shaft Angular Velocity |
| $\ddot{\vartheta}$[rad/$s^2$] | Shaft Angular Acceleration |
| $n$ | Gear Reduction Ratio |
| $I_l$[kg.$m^2$] | Load Inertia |
| $I_R$[kg.$m^2$] | Rotor Inertia |

$$\sum F_x = ma \to F_1 + F_1 + F_1 - mg\sin(\alpha) - F_d = ma = m\ddot{x} \tag{A.3}$$

$$\sum M_Y = I_{yy}\ddot{\phi} \to \frac{\sqrt{3}}{2}F_3 L\cos(\theta_3) - \frac{\sqrt{3}}{2}F_2 L\cos(\theta_2) = I_{yy}\ddot{\phi} \tag{A.4}$$

$$\sum M_z = I_{zz}\ddot{\psi} \to \frac{1}{2}F_3 L\cos(\theta_3) + \frac{1}{2}F_2 L\cos(\theta_2) - F_1 L\cos(\theta_1) - mg\cos(\alpha)\,L\sin(\theta_1) = I_{zz}\ddot{\psi} \tag{A.5}$$

The parameters in Eqs. (A.3)-(A.5) are listed in table A.2. The central processor is desired to remain at the center of pipe, hence, the arm angles ($\theta_i, i = 1,2,3$) are equal to each other and we have:

$$\cos(\theta_i) = \frac{D}{2L} \tag{A6}$$

Where $D$ is the pipe diamter and variable.

**Table A.2**

Parameters in Robot Dynamical Equations

| Parameter | Description | Value [unit] |
|---|---|---|
| $\theta_1$ | Arm #1 Angle | Variable [rad] |
| $\theta_2$ | Arm #2 Angle | Variable [rad] |
| $\theta_3$ | Arm #3 Angle | Variable [rad] |
| $D$ | Pipe Diameter | Variable [cm] |
| $L$ | Arms Length | 17 [$cm$] |
| $R$ | Wheel Radius | 5 [$cm$] |
| $m$ | Robot Mass | 2.23 [$kg$] |
| $I_{yy}$ | Robot's Moment of Inertia Around y-axis | 0.0126 [$kg.m^2$] |
| $I_{zz}$ | Robot's Moment of Inertia Around z-axis | 0.0093 [$kg.m^2$] |
| $\phi$ | Robot Rotation Angle Around y-axis | Variable [rad] |
| $\psi$ | Robot Rotation Angle Around z-axis | Variable [rad] |
| $F_1$ | Traction Force of Gear-motor #1 | Variable [N] |
| $F_2$ | Traction Force of Gear-motor #2 | Variable [N] |
| $F_3$ | Traction Force of Gear-motor #3 | Variable [N] |
| $F_d$ | Drag force | Variable [N] |

**Apprendix B.**

To design the LQR controller, considering the stabilizing states, $x_s = [\phi \quad \dot{\phi} \quad \psi \quad \dot{\psi}]^T$, Eq. (A.4) and Eq. (A.5) are linearized around the equilibrium point, $x_s^e = [0 \quad 0 \quad 0 \quad 0]^T$ with Taylor expansion and neglect higher order terms. The resulting equations and construct the system's auxiliary matrices. We have:

$$x_s = \begin{bmatrix} x_1 = \phi \\ x_2 = \dot{\phi} \\ x_3 = \psi \\ x_4 = \dot{\psi} \end{bmatrix} \tag{B.1}$$

$$u = \begin{bmatrix} \tau_1 = u_1 \\ \tau_2 = u_2 \\ \tau_3 = u_3 \end{bmatrix} \tag{B.2}$$

Hence, we can write:

$$\dot{x}_s = \begin{bmatrix} x_2 \\ \frac{1}{RI_{yy}}[\frac{\sqrt{3}}{2}\tau_3 L\cos(\theta_3) - \frac{\sqrt{3}}{2}\tau_2 L\cos(\theta_2)] \\ x_4 \\ \frac{1}{I_{zz}}[\frac{1}{2R}\tau_3 L\cos(\theta_3) + \frac{1}{2R}\tau_2 L\cos(\theta_2) - \frac{1}{R}\tau_1 L\cos(\theta_1) - mg\sin\alpha\sin\theta_1] \end{bmatrix} = \boldsymbol{F} = \begin{bmatrix} F_1 \\ F_2 \\ F_3 \\ F_4 \end{bmatrix} \tag{B.3}$$

$$A = \begin{bmatrix} \frac{\partial F_1}{\partial x_1} & \frac{\partial F_1}{\partial x_2} & \frac{\partial F_1}{\partial x_3} & \frac{\partial F_1}{\partial x_4} \\ \frac{\partial F_2}{\partial x_1} & \frac{\partial F_2}{\partial x_2} & \frac{\partial F_2}{\partial x_3} & \frac{\partial F_2}{\partial x_4} \\ \frac{\partial F_3}{\partial x_1} & \frac{\partial F_3}{\partial x_2} & \frac{\partial F_3}{\partial x_3} & \frac{\partial F_3}{\partial x_4} \\ \frac{\partial F_4}{\partial x_1} & \frac{\partial F_4}{\partial x_2} & \frac{\partial F_4}{\partial x_3} & \frac{\partial F_4}{\partial x_4} \end{bmatrix} = \begin{bmatrix} 0 & 1 & 0 & 0 \\ 0 & 0 & 0 & 0 \\ 0 & 0 & 0 & 1 \\ 0 & 0 & 0 & 0 \end{bmatrix} \tag{B.4}$$

$$B = \begin{bmatrix} \frac{\partial F_1}{\partial u_1} & \frac{\partial F_1}{\partial u_2} & \frac{\partial F_1}{\partial u_3} \\ \frac{\partial F_2}{\partial u_1} & \frac{\partial F_2}{\partial u_2} & \frac{\partial F_2}{\partial u_3} \\ \frac{\partial F_3}{\partial u_1} & \frac{\partial F_3}{\partial u_2} & \frac{\partial F_3}{\partial u_3} \\ \frac{\partial F_4}{\partial u_1} & \frac{\partial F_4}{\partial u_2} & \frac{\partial F_4}{\partial u_3} \end{bmatrix} = \begin{bmatrix} 0 & 0 & 0 \\ 0 & -\frac{\sqrt{3}D}{4RI_{yy}} & \frac{\sqrt{3}D}{4RI_{yy}} \\ 0 & 0 & 0 \\ -\frac{D}{2RI_{zz}} & \frac{D}{4RI_{zz}} & \frac{D}{4RI_{zz}} \end{bmatrix} \tag{B.5}$$

Pipe diameter, $D$, is genreally variable and since in our experiments, the pipe diameter is $\approx 18$ cm, we have $B = \begin{bmatrix} 0 & 0 & 0 \\ 0 & -123.72 & 123.72 \\ 0 & 0 & 0 \\ -193.55 & 96.77 & 96.77 \end{bmatrix}$. Based on the motion sensors in the robot, output matrix in state space representation,

$C = \begin{bmatrix} 0 & 1 & 0 & 0 \\ 0 & 0 & 1 & 0 \end{bmatrix}$ The state space representation of the system in terms of stabilization is written as:

$$\dot{x}_s = Ax_s + Bu$$
$$y = Cx_s$$

(B.6)

As for the LQR controller, a cost function, $J(K)$, is defined that is written as:

$$J(K) = \frac{1}{2}\int_0^{\infty}[x_s^T(t)Qx_s(t) + u(t)^T Ru(t)]dt \tag{B.7}$$

where $Q$ is the nonnegative definite matrix that weights state vector, $Q = \begin{bmatrix} 200 & 0 & 0 & 0 \\ 0 & 10 & 0 & 0 \\ 0 & 0 & 200 & 0 \\ 0 & 0 & 0 & 10 \end{bmatrix}$, and $R$ The positive-definite matrix that weights the input vector and $R = \begin{bmatrix} 1 & 0 & 0 \\ 0 & 1 & 0 \\ 0 & 0 & 1 \end{bmatrix}$. To minimize the cost function, $K$, gain matrix is computed with:

$$K = R^{-1}B^T P \tag{B.8}$$

$P$ in Eq. (B.8) is computed with algebraic Ricatti equation:

$$-PA - A^T P - Q + PBR^{-1}B^T P = 0 \tag{B.9}$$

We have $K = \begin{bmatrix} -4.92 & -1.12 & -13.26 & -2.98 \\ -9.37 & -2.11 & 3.48 & 0.78 \\ -9.37 & -2.11 & 3.48 & 0.78 \end{bmatrix}$ And the control input of the LQR controller is computed as:

$$u = -Kx_s \tag{B.10}$$

As for velocity controllers, we considered three PID controllers that each of them, controls the velocity of one wheel. The parameters of the PID controllers are listed in the Table B.1.

**Table B.1**

PID Parameters

| **Parameter** | **Description** | **Value** |
|---|---|---|
| $K_P$ | Proportional Gain | 8.7313 |
| $K_I$ | Integral Gain | 322.4160 |
| $K_D$ | Derivative Gain | 0.0073 |

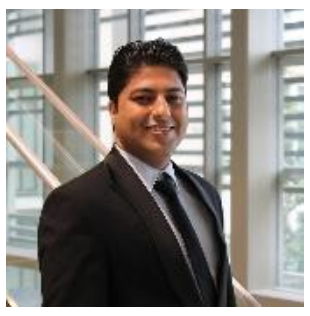

Saber Kazeminasab is a PhD student in the Department of Electrical and Computer Engineering, Texas A&M University, College Station, TX, USA. He received the B.Sc. degree from the Iran University of Science and Technology, Tehran, Iran, in 2014, in mechanical engineering, and the M.Sc. degree from the University of Tehran, Tehran, Iran, in 2017 in mechatronics engineering. His research interests include mechatronics, robotics, control theories, mechanism design, and actuator design.

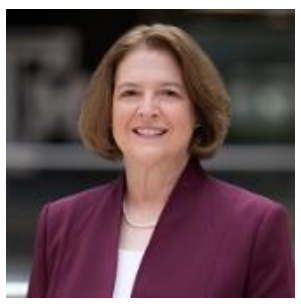

M. Katherine Banks is a professor of Civil engineering department and currently Vice Chancellor of Engineering for The Texas A&M University System and Dean of the Texas A&M University College of Engineering. She is an Elected Fellow of the American Society of Civil Engineers, was elected in 2014 to the National Academy of Engineering, and was formerly the Jack and Kay Hockema Professor at Purdue University. Her research interests include applied microbial systems, biofilm processes, wastewater treatment and reuse, and phytoremediation bioremediation. She received her Ph.D. in 1989 from Duke University. At Texas A&M, she helped establish the EnMed program (led by Roderic Pettigrew, Ph.D., M.D.), an innovative engineering medical school option created by Texas A&M University and Houston Methodist Hospital, designed to educate a new kind of physician who will create transformational technology for health care. She received her Bachelor of Science in Engineering from the University of Florida, Master of Science in Engineering from the University of North Carolina, and Doctorate of Philosophy in civil and environmental engineering from Duke University. Dr. Banks is the recipient of the American Society of Civil Engineers Petersen Outstanding Woman of the Year Award, American Society of Civil Engineers Rudolph Hering Medal, Purdue Faculty Scholar Award, Sloan Foundation Mentoring Fellowship and the American Association of University Women Fellowship. On February 13, 2019, she was named to the Board of Directors of Halliburton.